\documentclass{isprs} 
\usepackage{subfigure}
\usepackage{setspace}
\usepackage{geometry} 
\usepackage{epstopdf}
\usepackage[labelsep=period]{caption}  
\usepackage[british]{babel} 
\usepackage[hang]{footmisc}

\usepackage{caption}
\usepackage{amsmath}
\usepackage{amssymb}
\usepackage{todonotes}
\usepackage{graphicx}
\usepackage{algorithm}
\usepackage{algpseudocode}
\usepackage{subfigure}
\usepackage{booktabs}
\usepackage{multirow}
\usepackage{xcolor}
\definecolor{DarkGreen}{RGB}{0,128,0}
\usepackage{colortbl}
\usepackage{tikz}        
\usepackage{pgfplots}
\usepgfplotslibrary{colorbrewer}
\usepackage{url}

\usepackage{placeins}
\usepackage[authoryear]{natbib}
\usepackage[acronym]{glossaries}
\makeglossaries
\newacronym{gs}{GS}{Gaussian Splatting}
\newacronym{3dgs}{3DGS}{3D Gaussian Splatting}
\newacronym{2dgs}{2DGS}{2D Gaussian Splatting}
\newacronym{sfm}{SfM}{Structure-from-Motion}
\newacronym{nerf}{NeRF}{Neural Radiance Fields}
\newacronym{ssim}{SSIM}{structural similarity}
\newacronym{nvs}{NVS}{novel view synthesis}
\newacronym{mvs}{MVS}{Multi-View Stereo}
\newacronym{tnt}{TnT}{Tanks and Temples}
\newacronym{lod}{LoD}{Level of Detail}
\newacronym{lidar}{LiDAR}{Light Detection and Ranging}
\newacronym{gs4b}{GS4B}{Gaussian4Buildings}

\begin{document}

\title{GS4Buildings: Prior-Guided Gaussian Splatting for 3D Building Reconstruction}
\date{}

\author{
 Qilin Zhang, Olaf Wysocki, Boris Jutzi}

\address{
	Photogrammetry and Remote Sensing, TUM School of Engineering and Design, Technical University of Munich (TUM), \\Munich, Germany - (qilin.zhang, olaf.wysocki, boris.jutzi)@tum.de\\
}


\abstract{
Recent advances in \gls{gs} have demonstrated its effectiveness in photo-realistic rendering and 3D reconstruction. 
Among these, \gls{2dgs} is particularly suitable for surface reconstruction due to its flattened Gaussian representation and integrated normal regularization. 
However, its performance often degrades in large-scale and complex urban scenes with frequent occlusions, leading to incomplete building reconstructions.
We propose GS4Buildings, a novel prior-guided Gaussian Splatting method leveraging the ubiquity of semantic 3D building models for robust and scalable building surface reconstruction. 
Instead of relying on traditional \gls{sfm} pipelines, GS4Buildings initializes Gaussians directly from low-level \gls{lod}2 semantic 3D building models. 
Moreover, we generate prior depth and normal maps from the planar building geometry and incorporate them into the optimization process, providing strong geometric guidance for surface consistency and structural accuracy. 
We also introduce an optional building-focused mode that limits reconstruction to building regions, achieving a 71.8\% reduction in Gaussian primitives and enabling a more efficient and compact representation. 
Experiments on urban datasets demonstrate that GS4Buildings improves reconstruction completeness by 20.5\% and geometric accuracy by 32.8\%. 
These results highlight the potential of semantic building model integration to advance GS-based reconstruction toward real-world urban applications such as smart cities and digital twins. 
Our project is available: \url{https://github.com/zqlin0521/GS4Buildings}.
}

\keywords{Gaussian Splatting, Semantic 3D Building Models, CityGML, LoD2, Geometric Supervision, 3D Reconstruction}

\maketitle

\section{Introduction}\label{Introduction}
\glsresetall
\begin{sloppypar}
Novel view synthesis (NVS) and 3D reconstruction have become fundamental techniques in computer vision and photogrammetry, driven by growing demands in virtual reality, urban planning, and digital twin applications. 
With advances in computer graphics, \gls{gs} has recently emerged as a powerful approach, offering state-of-the-art performance in photo-realistic rendering and high-fidelity scene reconstruction. 
\gls{3dgs}~\citep{kerbl20233d} represents scenes using anisotropic 3D Gaussian primitives, whose positions, orientations, and appearance parameters are jointly optimized from multi-view image data. 
Leveraging an efficient tile-based rasterization technique, \gls{3dgs} enables real-time rendering while maintaining high visual fidelity. 
To better support surface reconstruction tasks, \gls{2dgs}~\citep{huang20242d} extends this framework by introducing flattened Gaussian representations and incorporating normal-based regularization.

\begin{figure}[t]
    \centering
    \subfigure[MVS]{
        \includegraphics[width=0.14\textwidth]{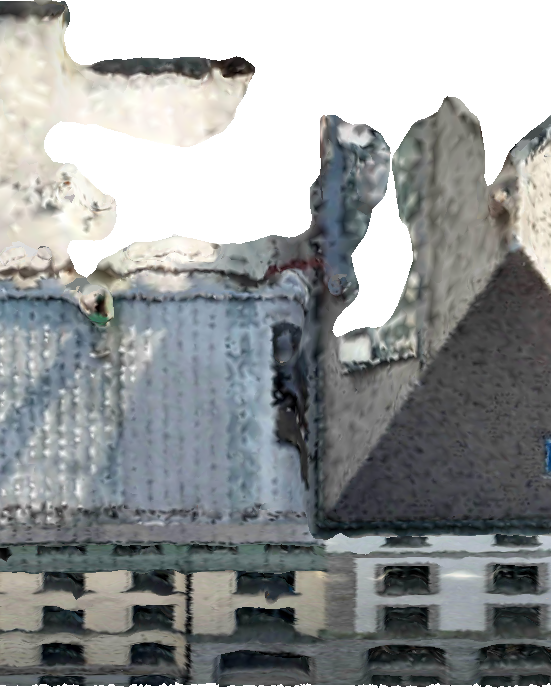}
    }
    \subfigure[2DGS]{
        \includegraphics[width=0.14\textwidth]{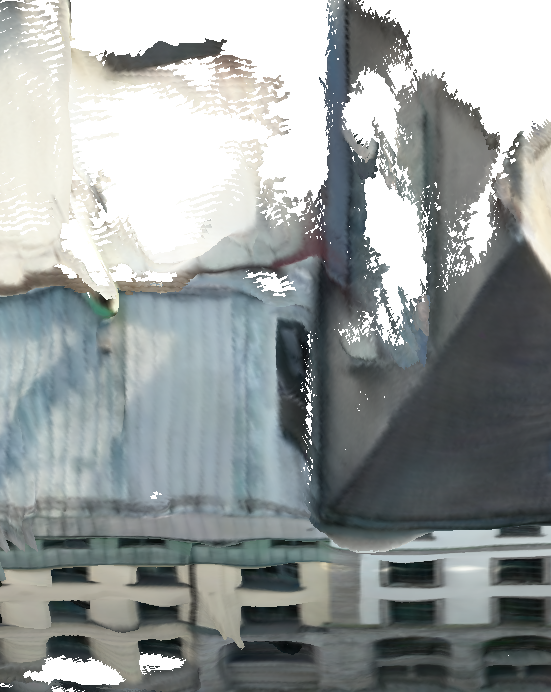}
    }
    \subfigure[GS4B (Ours)]{
        \includegraphics[width=0.14\textwidth]{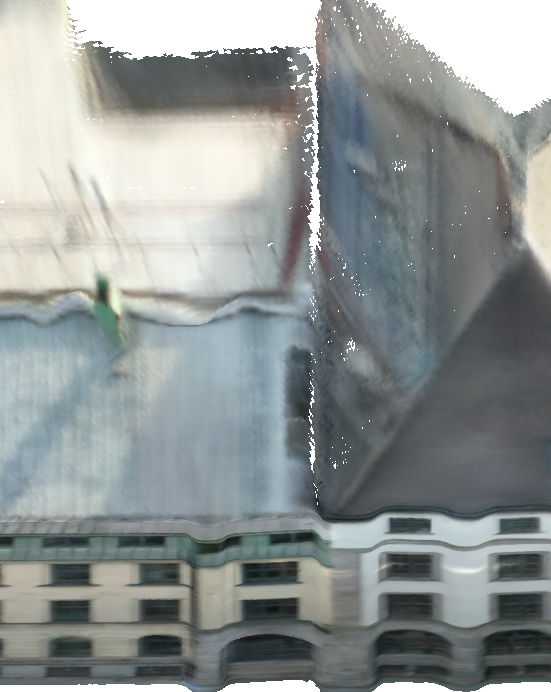}
    }
    \caption{Unlike (a) traditional \gls{mvs} and (b) vanilla \gls{2dgs},  both render incomplete building surfaces, (c) our Gaussian4Buildings (GS4B) method reconstructs complete building shape.}
    \label{fig:intro_comparison}
\end{figure}

While recent innovations have improved \gls{gs} performance, existing methods still struggle with urban-scale building reconstruction. 
As illustrated in Figure~\ref{fig:intro_comparison}, \gls{2dgs} often fails to recover complete building surfaces under occlusions or limited viewpoint coverage. 
Traditional \gls{mvs} pipelines similarly struggle in textureless or repetitive regions, where reliable feature matching is difficult. 
These challenges, observed in both \gls{gs} and \gls{mvs} approaches, limit their suitability for urban reconstruction tasks that demand high geometric completeness. 
Concurrently, the development of smart cities has led to the widespread availability of semantic 3D building models with more than 215 million open source building models worldwide~\citep{wysocki2024reviewing}. 
These models are lightweight, volumetric, and typically represented in boundary representation (B-Rep) geometry. 
They encode reliable geometric information such as roof structures, wall orientations, and building footprints, 
and are generally constructed under a planar surface assumption, making them well-suited as geometric priors for surface reconstruction. 
Although earlier studies have explored the integration of \gls{lod}2 building models with point clouds~\citep{wysocki2023scan2lod3,helmutMayerLoD3}, the building models' potential for guiding \gls{gs}-based reconstruction remains largely underexplored. 

To fill this gap, we propose GS4Buildings, a prior-guided \gls{gs} framework that enables robust and complete reconstruction of buildings under challenging urban conditions, including large occlusions, complex architectural geometry, and sparse camera viewpoints. 
Instead of relying on \gls{sfm} pipelines for initialization, our method samples 3D points directly from the \gls{lod}2 building model and filters them based on multi-view visibility using known camera intrinsics and extrinsics. 
The resulting point cloud is used to initialize the Gaussian splats in a geometry-consistent manner. 
In addition, depth and normal maps projected from the \gls{lod}2 geometry are integrated into the optimization process, providing surface-aware constraints that guide the Gaussians toward architecturally consistent reconstructions. 
We also introduce an optional building-focused mode that reduces memory and computation overhead while preserving reconstruction quality. 
This makes GS4Buildings practical for large-scale applications with constrained resources. 
\begin{figure*}[ht!]
    \centering
    \includegraphics[width=\textwidth]{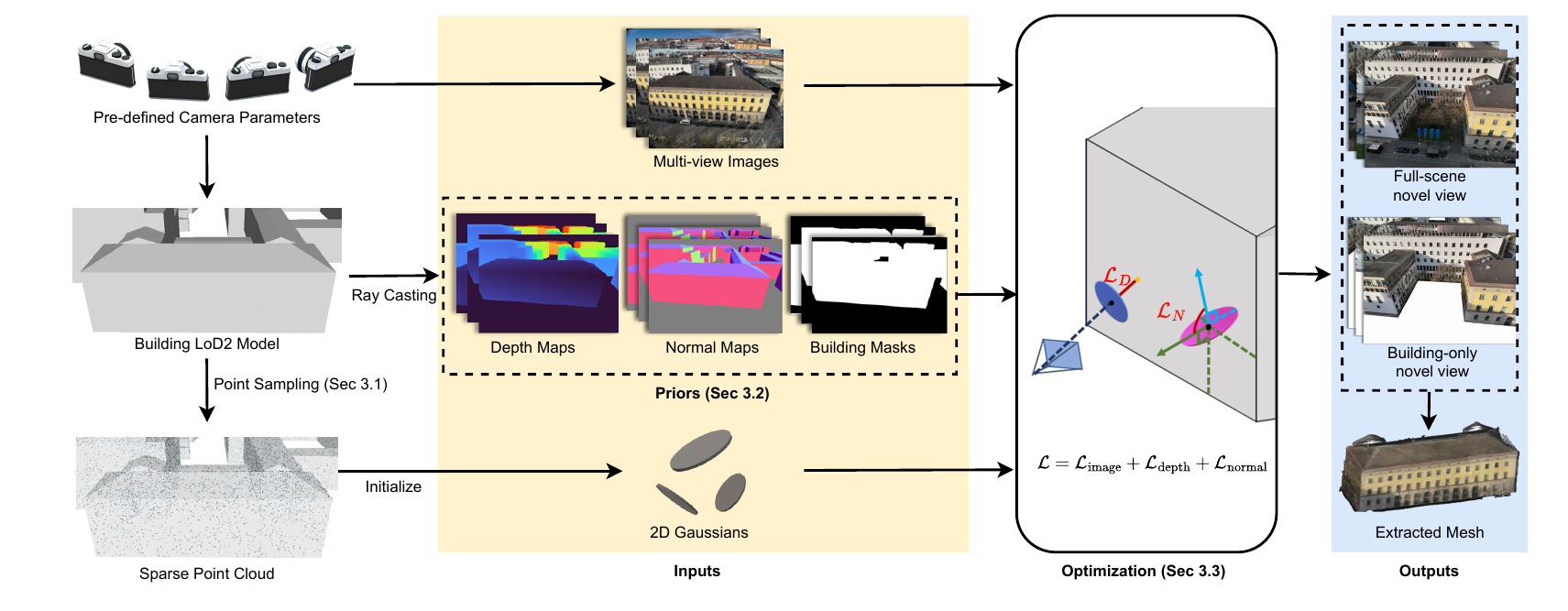}
    \caption{Overview of the proposed GS4Buildings framework. 
    Given a \gls{lod}2 building model and camera parameters, our method samples 3D points from the mesh surface and filters them by multi-view visibility to initialize 2D Gaussians. 
    Prior depth and normal maps projected from the \gls{lod}2 geometry are incorporated during optimization to guide the reconstruction. }
    \label{fig:overview}
\end{figure*}

In summary, the main contributions of this work are as follows:
\begin{itemize}
    \item We present GS4Buildings, a \gls{sfm}-free \gls{gs} framework for building reconstruction, which directly leverages \gls{lod}2 building models and camera parameters to initialize geometry-aware Gaussian splats.
    \item We enhance the optimization of \gls{2dgs} by incorporating depth and normal priors derived from the \gls{lod}2 geometry, enabling more complete and accurate reconstruction of building surfaces, particularly under occlusions and limited viewpoint conditions. 
    \item We introduce an optional building-focused reconstruction mode that concentrates computation on architecturally relevant regions, offering improved efficiency while maintaining reconstruction quality. 
\end{itemize}

\end{sloppypar}
\section{Related Work}\label{Related Work}
\begin{sloppypar}
This section reviews geometry-aware enhancements for \gls{gs} and recent advances in building reconstruction, the main application focus of this work. 
\paragraph{Gaussian Splatting for 3D Reconstruction }
\gls{gs} has recently gained significant attention as a powerful scene representation technique for novel view synthesis and real-time rendering. 
While \gls{3dgs}~\citep{kerbl20233d} achieves impressive visual quality, it faces several challenges when applied to 3D reconstruction, especially surface reconstruction. 
The volumetric radiance representation of \gls{3dgs} is incompatible with the thin, structured nature of real-world surfaces, and its rasterization suffers from multi-view inconsistencies, leading to noisy or incomplete reconstructions~\citep{huang20242d}. 

To address the limitations of \gls{3dgs}, recent works have proposed incorporating geometric priors into the \gls{gs} framework. 
One common strategy is to integrate monocular depth estimation into the optimization process. For instance, DN-Gaussian~\citep{li2024dngaussian} and CDGS~\citep{zhang2025cdgs} use predicted depth maps to provide supervision for Gaussian placement and refinement. 
These methods apply depth-aware losses, such as gradient- or edge-sensitive regularization, to improve geometric consistency. 
Other works enhance surface modeling by adjusting the Gaussian shape and orientation to better reflect local geometry. 
2DGS~\citep{huang20242d} represents each Gaussian as an elliptical disk embedded in a local tangent plane, thereby aligning the primitive with the underlying surface. 
DN-Splatter~\citep{turkulainen2024dn} encourages Gaussians to take on a disc-like shape during optimization, with their smallest scaling axis aligned to the surface normal, guided by depth and normal priors. 
While these geometry-aware designs improve reconstruction quality under ideal conditions, their reliance on image-based estimations and \gls{sfm}-based initialization still limits their effectiveness in complex urban environments. 
For urban building reconstruction in particular, directly leveraging existing 3D models of buildings as geometric priors holds strong potential for improving completeness and accuracy.

\paragraph{Building Reconstruction }
Traditional methods for building reconstruction primarily rely on \gls{sfm} combined with \gls{mvs}~\citep{schonberger2016structure}, or on high-precision terrestrial or airborne \gls{lidar} scanning~\citep{HAALA2010570}. 
\gls{sfm}-\gls{mvs} approaches reconstruct dense point clouds or surface meshes from multi-view images and perform well in well-conditioned scenarios, but often struggle with occlusions, repetitive structures, or insufficient viewpoint coverage. 
\gls{lidar}-based techniques provide accurate geometric measurements and are more robust to such challenges, yet are limited in texture detail and appearance quality, reducing their effectiveness in appearance-aware modeling tasks. 

In parallel with image- and point-based reconstruction approaches, structured urban models have become an important component in large-scale 3D city modeling. 
Among them, the CityGML standard defining the \gls{lod} framework~\citep{grogerOGCCityGeography2012} offers standardized, hierarchical representations of building geometry, ranging from simple building blocks to highly detailed architectural elements. 
Currently, low-level \gls{lod}1 represented by cuboid-like objects and \gls{lod}2 with complex roof shapes and simplified facades can be automatically reconstructed given footprints and aerial observations.
This trend enabled a wide adoption of such models worldwide, totalling more than 215 million open data building models in countries such as Poland, the Netherlands, or Japan~\citep{wysocki2024reviewing}. 
Owing to their lightweight representation, structured geometry, and inherent planarity, \gls{lod}2 models can serve as valuable geometric priors for aligning and refining the reconstruction process.

Recent advances in urban scene representation have explored both implicit methods such as \gls{nerf}~\citep{mildenhall2021nerf} and explicit representations such as \gls{gs}~\citep{kerbl20233d}. 
While \gls{nerf} excels at photorealistic view synthesis, \gls{gs} offers better scalability and efficiency. 
To extend its use in geometry-oriented tasks, recent works have focused on extracting more generalizable 3D models, such as surface meshes, from GS outputs. 
For example, 2DGS~\citep{huang20242d} improves mesh reconstruction quality, and Gaussian Building Mesh~\citep{gao2024gaussian} further adapts \gls{2dgs} to structured architectural modeling using semantic masks. 
To the best of our knowledge, 
we are the first to incorporate structured geometric models, such as \gls{lod}2, into \gls{gs}-based reconstruction pipelines for urban buildings, 
enabling both improved initialization and geometry-aware optimization.

\end{sloppypar}

\section{Methodology}
\label{sec:method}
\begin{sloppypar}
Our GS4B method enhances \gls{2dgs} by incorporating geometric priors derived from semantic \gls{lod}2 building models. As illustrated in Figure~\ref{fig:overview}, the proposed framework begins with a geometry-consistent Gaussian initialization guided by the \gls{lod}2 mesh and camera poses (Section~\ref{sec:init}). We then generate dense depth and normal priors for each view using a raycasting-based approach (Section~\ref{sec:prior-generation}). These priors are incorporated into a prior-guided optimization scheme that extends \gls{2dgs} training process (Section~\ref{sec:2dgs_with_priors}). Finally, we extract a surface mesh from the optimized Gaussians using volumetric fusion.
\subsection{Prior-Based Gaussian Initialization}
\label{sec:init}
With the advancement of sensor technology and lower data acquisition costs, accurate camera parameters are now accessible without relying on image-based \gls{sfm} methods.  
Leveraging this, we adopt a prior-based initialization strategy guided by the \gls{lod}2 building mesh and known camera poses. 
Unlike conventional \gls{gs} pipelines such as \gls{2dgs}, which rely on \gls{sfm} to reconstruct sparse geometry, our method directly samples a structurally reliable point cloud from the \gls{lod}2 mesh, enabling geometry-consistent initialization of 2D Gaussian primitives. 
We begin by uniformly sampling 3D points on the mesh surface using face-area-weighted random sampling~\citep{trimesh}. 
For each sampled point $\mathbf{p}_i \in \mathbb{R}^3$ and camera $j$ with position $\mathbf{c}_j$, we define the expected depth as:
\begin{equation}
    d_{i,j}^{\text{exp}} = \| \mathbf{p}_i - \mathbf{c}_j \|_2.
\end{equation}
Let $d_{i,j}^{\text{int}}$ be the depth of the first ray–mesh intersection along the viewing direction from $\mathbf{c}_j$ to $\mathbf{p}_i$. The point is considered visible in view $j$ if:
\begin{equation}
    v_{i,j} = 
    \begin{cases}
        1, & \text{if } \left| d_{i,j}^{\text{int}} - d_{i,j}^{\text{exp}} \right| < \varepsilon \\
        0, & \text{otherwise}
    \end{cases}
\end{equation}
where $\varepsilon$ is a distance threshold (e.g., 5 cm). A point is retained if it is visible in at least $k$ views:
\begin{equation}
    \sum_{j=1}^M v_{i,j} \geq k.
\end{equation}
For each retained point, we store its 3D coordinates along with its 2D projections $\pi_j(\mathbf{p}_i)$ in all observing views where $v_{i,j} = 1$. 
The resulting data is organized in a format compatible with standard \gls{gs} pipelines, enabling seamless Gaussian initialization and consistent association with image projections. 
Compared to \gls{sfm}-based initialization, our strategy directly samples from the structured \gls{lod}2 geometry, enabling efficient, density-controllable point selection that focuses on geometrically meaningful building regions. 
This design also avoids the potential limitations of \gls{sfm} in urban scenarios, such as failure cases or noisy reconstructions. 

\subsection{Prior Generation from Building Models}
\label{sec:prior-generation}
To guide the optimization of Gaussian primitives toward more accurate and complete surface reconstruction, we generate per-view depth and normal priors from the \gls{lod}2 building geometry. 
These priors provide reliable geometric supervision, especially in regions where image-based cues are unreliable due to occlusion or limited view coverage.

Specifically, we transform the \gls{lod}2 mesh $\mathcal{M}_\text{raw}$ into the global scene coordinate system defined by the camera poses, resulting in a consistent mesh representation $\mathcal{M}$. 
Given this aligned mesh and a set of calibrated cameras with intrinsics $\mathbf{K}_j$ and extrinsics $\mathbf{T}_j$, we adopt a raycasting-based approach~\citep{zhou2018open3d} to synthesize dense, view-aligned geometric priors. For each camera view $j$, the raycasting process computes:
\begin{equation}
    \mathbf{D}_j, \mathbf{N}_j = \mathcal{R}(\mathcal{M}, \mathbf{K}_j, \mathbf{T}_j),
\end{equation}
where $\mathbf{D}_j \in \mathbb{R}^{H \times W}$ is the depth map and $\mathbf{N}_j \in \mathbb{R}^{H \times W \times 3}$ is the corresponding normal map. During raycasting, we also record a binary visibility mask $\mathbf{M}_j \in \{0,1\}^{H \times W}$ that indicates which pixels result in valid mesh intersections; this mask is later used to exclude invalid pixels during loss computation.

Unlike image-based estimations affected by occlusions, lighting, or network uncertainty, prior-guided supervision offers consistent, occlusion-free geometric guidance derived from the building structure, while requiring less computational effort. 
These priors are later integrated into the optimization process.
\subsection{\gls{2dgs} with Prior-Guided Optimization}
\label{sec:2dgs_with_priors}
\begin{figure}[ht]
    \centering

    \subfigure[Depth supervision $\mathcal{L}_{\text{d\_b}}$]{
        \includegraphics[width=0.47\linewidth]{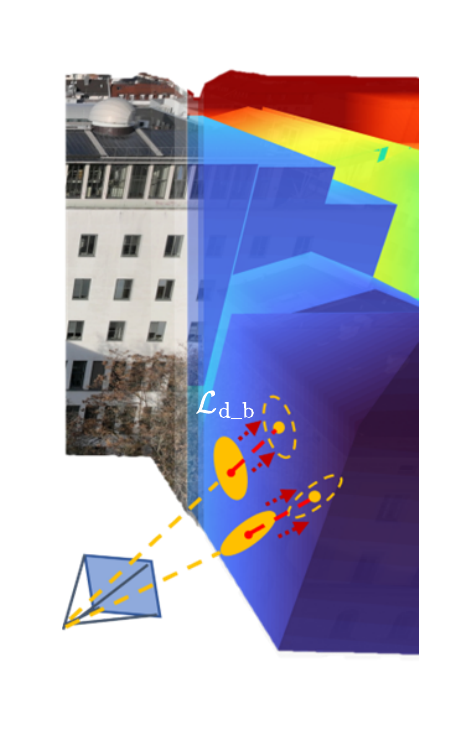}
    }
    \hfill
    \subfigure[Normal supervision $\mathcal{L}_{\text{n\_b}}$]{
        \includegraphics[width=0.47\linewidth]{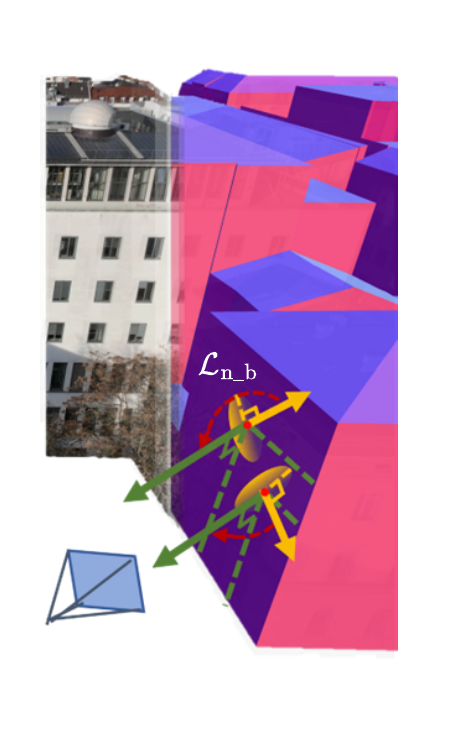}
    }
    \caption{
        Prior-guided supervision: (a) Depth priors align splats with surface geometry; (b) Normal priors ensure consistent orientation across views.
    }
    \label{fig:method}
\end{figure}

Building on the prior-guided initialization and dense priors, we extend the \gls{2dgs} optimization pipeline with structured supervision from building models. 
\paragraph{\gls{2dgs} Formulation}
\gls{2dgs}~\citep{huang20242d} represents a scene using elliptical 2D Gaussian splats, each defined by a center point $\mathbf{p}_k \in \mathbb{R}^3$, two tangent vectors $\mathbf{t}_u, \mathbf{t}_v \in \mathbb{R}^3$, and anisotropic scaling factors $(s_u, s_v)$ within the tangent plane:
\begin{equation}
    P(u, v) = \mathbf{p}_k + s_u u \mathbf{t}_u + s_v v \mathbf{t}_v.
\end{equation}
Each splat is rendered into screen space using an explicit ray–splat intersection strategy. Its density is modeled as a Gaussian:
\begin{equation}
    \mathcal{G}(u, v) = \exp\left( -\frac{u^2 + v^2}{2} \right),
\end{equation}
and the final pixel color $\mathbf{c}(\mathbf{x})$ is obtained through alpha compositing:
\begin{equation}
    \mathbf{c}(\mathbf{x}) = \sum_{i=1}^N \alpha_i \mathbf{c}_i \mathcal{G}_i(\mathbf{u}(\mathbf{x})) \prod_{j=1}^{i-1} (1 - \alpha_j \mathcal{G}_j(\mathbf{u}(\mathbf{x}))).
\end{equation}
To encourage geometric consistency, 2DGS introduces two regularization losses: a depth distortion loss $\mathcal{L}_d$ and a normal consistency loss $\mathcal{L}_n$:
\begin{equation}
    \mathcal{L}_n = \sum_i \omega_i \left( 1 - \mathbf{n}_i^\top \mathbf{N}(\mathbf{x}_i) \right),
\end{equation}
where $\mathbf{n}_i$ is the Gaussian's surface normal and $\mathbf{N}(\mathbf{x}_i)$ is the normal estimated from the rendered depth gradient. 
The original \gls{2dgs} training objective is:
\begin{equation}
    \mathcal{L}_{\text{2DGS}} = \mathcal{L}_c + \lambda_d \mathcal{L}_d + \lambda_n \mathcal{L}_n.
\end{equation}
\paragraph{Prior-Guided Optimization}
To further enhance building-oriented reconstruction, we introduce building-aware supervision derived from \gls{lod}2-based priors $(\mathbf{D}_j, \mathbf{N}_j)$ obtained via raycasting (see Section~\ref{sec:prior-generation}). 
Given a binary mask $\mathbf{M}_j$ indicating valid mesh intersections, we define two additional losses:
\begin{equation}
    \mathcal{L}_{\text{d\_b}} = \frac{1}{|\mathbf{M}_j|} \sum_{(x, y) \in \mathbf{M}_j} \left| \alpha \cdot \hat{\mathbf{D}}_j(x, y) - \mathbf{D}_j(x, y) \right|,
\end{equation}
\begin{equation}
    \mathcal{L}_{\text{n\_b}} = \frac{1}{|\mathbf{M}_j|} \sum_{(x, y) \in \mathbf{M}_j} \left( 1 - \langle \hat{\mathbf{N}}_j(x, y), \mathbf{N}_j(x, y) \rangle \right),
\end{equation}
where $\hat{\mathbf{D}}_j$ and $\hat{\mathbf{N}}_j$ are the rendered depth and normal maps from the current model, and $\alpha$ is a scale adjustment factor. Figure~\ref{fig:method} illustrates how these losses guide 2D Gaussian splats toward accurate surface reconstruction. 

We adopt a two-phase training strategy:
\begin{itemize}
    \item Phase 1: Emphasize $\mathcal{L}_{\text{d\_b}}$ and $\mathcal{L}_{\text{n\_b}}$ to enforce global geometric completeness and correctness using building priors.
    \item Phase 2: Gradually reduce prior-based loss weights while activating $\mathcal{L}_d$ and $\mathcal{L}_n$ to refine surface smoothness and alignment through local consistency.
\end{itemize}
The overall training objective becomes:
\begin{equation}
    \mathcal{L}_{\text{total}} = \mathcal{L}_c + \lambda_d \mathcal{L}_d + \lambda_n \mathcal{L}_n + \lambda_{\text{d\_b}} \mathcal{L}_{\text{d\_b}} + \lambda_{\text{n\_b}} \mathcal{L}_{\text{n\_b}},
\end{equation}
where all loss weights $\lambda$ are time-dependent and scheduled to balance prior guidance and visual fidelity throughout training. 

Additionally, we support two training modes. The \textit{building-only} mode restricts both view sampling and optimization to pixels within the building region (as defined by $\mathbf{M}_j$), effectively limiting the \gls{gs} reconstruction to the building area. 
The \textit{building-enhanced} mode retains full-scene training while applying additional supervision within building areas. 
To obtain a structured surface representation, we extract a mesh from the optimized Gaussians using TSDF fusion and Marching Cubes, following the 2DGS pipeline~\citep{huang20242d}. 
This yields a watertight mesh suitable for downstream geometric evaluation.

In summary, our framework combines prior-consistent initialization, occlusion-free geometric supervision, flexible training modes, and TSDF-based mesh extraction to enable accurate and complete 3D reconstruction in complex urban environments. 

\end{sloppypar}
\section{Experiments}
\label{Experiments}
\begin{sloppypar}
We evaluated our GS4B method in terms of NVS and 3D reconstruction quality, with comparisons to the original \gls{2dgs} and a conventional \gls{mvs} pipeline. 
This section outlines the experimental setup and presents qualitative and quantitative results. 

\subsection{Dataset and Evaluation Metrics}
\label{sec:data_eval}
We conducted experiments on the publicly available TUM2TWIN dataset~\citep{tum2twin}, which offers a comprehensive multi-modal capture of the central campus of the Technical University of Munich and its surrounding urban areas. 
For reconstruction input, we used UAV photographs and \gls{lod}2 building models, the latter serving as structured geometric priors for initialization and supervision. 
The UAV imagery collection~\citep{tum2twin_uav} includes 1,179 high-resolution photographs covering more than 70 buildings. 
To ensure diversity and coverage, we defined nine representative subsets, each focusing on a distinct building cluster and containing approximately 10–30 images. These subsets span a variety of building forms and urban complexities under real-world conditions.

For 2D photometric evaluation, we adopted three widely used metrics: Peak Signal-to-Noise Ratio (PSNR) for pixel-wise accuracy, Structural Similarity Index (SSIM) for structural fidelity, and Learned Perceptual Image Patch Similarity (LPIPS) for perceptual quality. 
We further assessed 3D reconstruction quality using two types of reference point clouds from the TUM2TWIN project. 
The first was laser-scanned point clouds, providing fine-grained geometric accuracy and acquired synchronously with the UAV imagery~\citep{tum2twin_uav}. 
The second was \gls{lod}3-derived point clouds, which are structurally coherent with the \gls{lod}2 models and served as a reference for assessing structural completeness. 

To evaluate geometric accuracy, we used two distance-based metrics: 
Chamfer Distance (CD), which measures global similarity by averaging bidirectional nearest-neighbor distances, 
and Multi-Scale Model-to-Model Cloud Comparison (M3C2)~\citep{lague2013accurate}, which captures orientation-aware surface deviations.  
For completeness evaluation, we employed two complementary strategies: 
threshold-based completeness, adapted from the Tanks and Temples benchmark~\citep{knapitsch2017tanks}, which calculates the fraction of ground-truth points within predefined distance thresholds (e.g., 0.1\,m to 0.5\,m); 
and voxel occupancy completeness (VOC),  inspired by the geometric completeness ~\citep{jager20233d}, which assesses volumetric coverage by comparing voxel occupancy between the reconstruction and the reference.  
A voxel is considered occupied if it contains at least a predefined number of points.

\subsection{Implementation Details}
Our implementation was based on \gls{2dgs} codebase~\citep{huang20242d}. 
We trained the model for 30{,}000 iterations across all experiments. 
Camera pose estimation and \gls{mvs}-based reconstruction results were generated using Pix4Dmatic~\citep{pix4dmatic} with default parameters. 
For initialization and prior generation, we employed \texttt{trimesh}~\citep{trimesh} for surface sampling and Open3D~\citep{zhou2018open3d} for raycasting and mesh operations. 
We followed the two-stage loss scheduling strategy described in Section~\ref{sec:2dgs_with_priors}, with all \gls{2dgs} hyperparameters kept unchanged for fair comparison. 
\end{sloppypar}
\subsection{Results}
\begin{figure}[ht]
    \centering
    \subfigure[GT]{
        \includegraphics[width=0.3\linewidth]{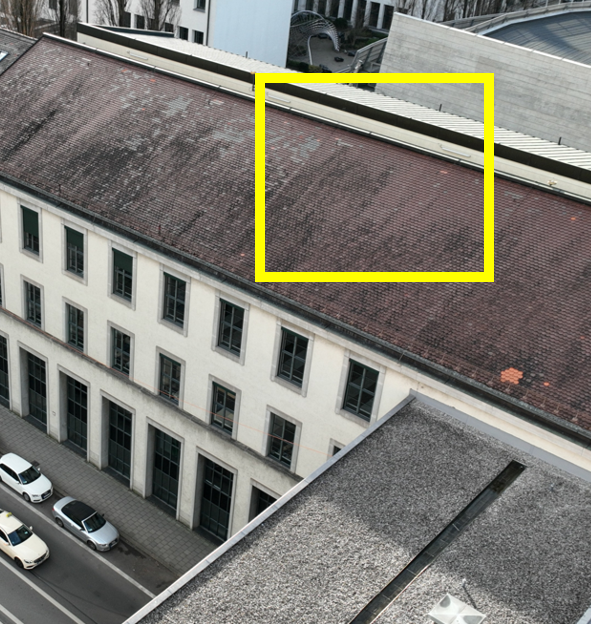}
    }
    \subfigure[2DGS]{
        \includegraphics[width=0.3\linewidth]{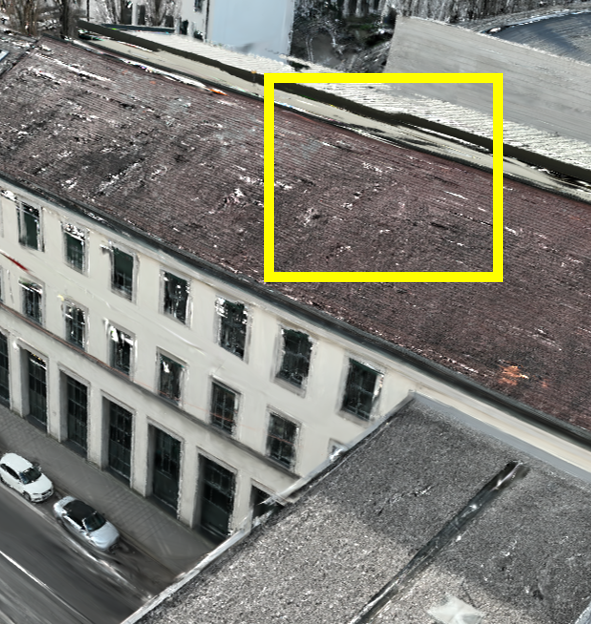}
    }
    \hfill
    \subfigure[GS4B (ours)]{
        \includegraphics[width=0.3\linewidth]{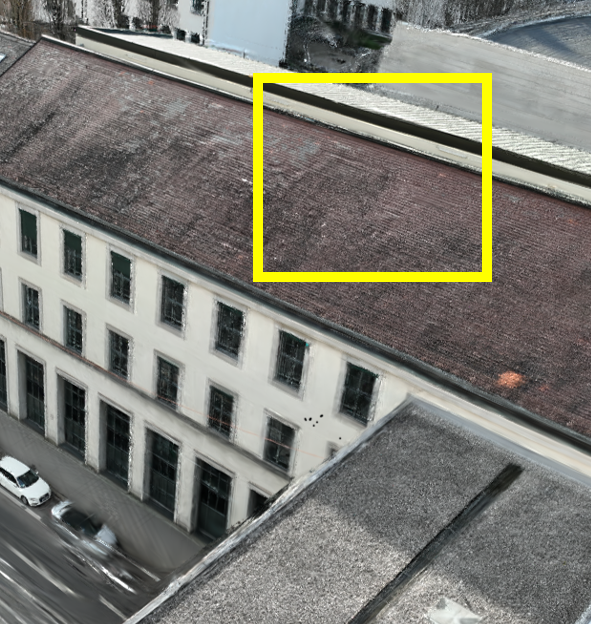}
    }
    \subfigure[GT]{
        \includegraphics[width=0.3\linewidth]{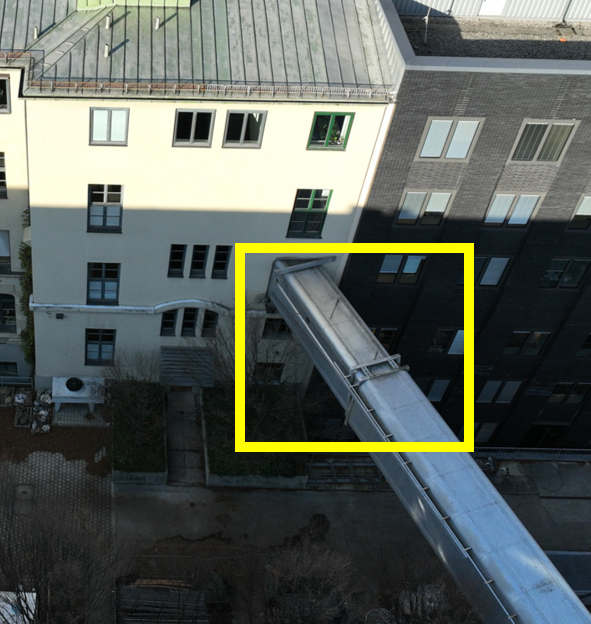}
    }
    \subfigure[2DGS]{
        \includegraphics[width=0.3\linewidth]{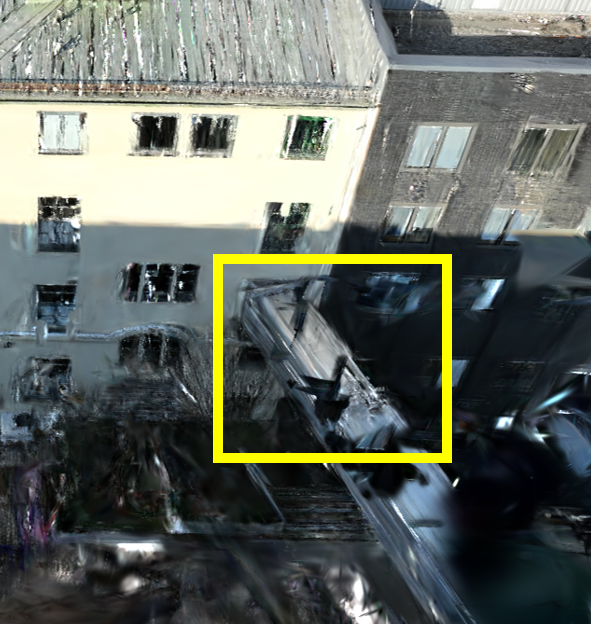}
    }
    \hfill
    \subfigure[GS4B (ours)]{
        \includegraphics[width=0.3\linewidth]{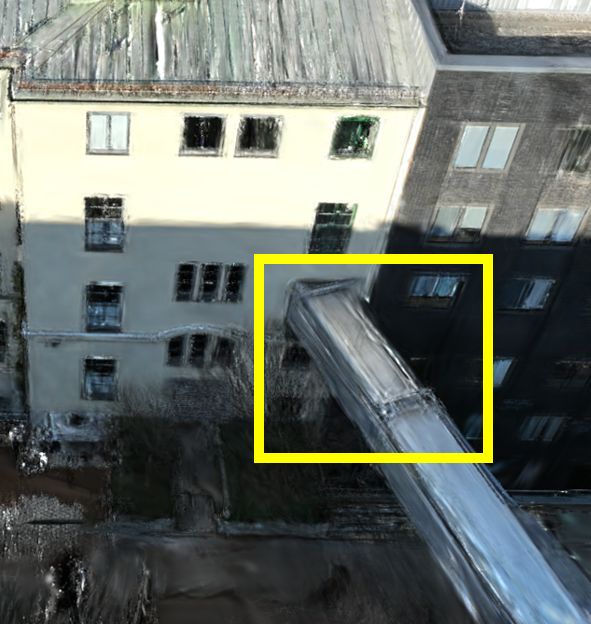}
    }
    \caption{
        2D visual comparison of NVS results on representative building scenes. 
        Our method achieves better visual quality 
        than \gls{2dgs} in challenging areas, such as textureless rooftops (b, c) and sparsely observed regions (e, f).
    }
    \label{fig:nvs_qualitative}
\end{figure}

\begin{sloppypar}
We evaluated our method against two baselines: \gls{2dgs}~\citep{huang20242d} and the \gls{mvs} pipeline from Pix4Dmatic. 
Evaluation was conducted from two perspectives: (i) 2D NVS quality, compared with \gls{2dgs} using perceptual and pixel-wise metrics; and (ii) 3D reconstruction quality, compared with both baselines in terms of accuracy and completeness. 
This dual-perspective analysis demonstrated the effectiveness of integrating building-aware priors into the \gls{gs} framework. 
\paragraph{Novel View Synthesis}
We assessed the effectiveness of our method in NVS through both qualitative and quantitative analyses. 
To ensure a fair comparison of scene-level reconstruction capabilities, we compared the results of our method in \textit{building-enhanced} mode against the original \gls{2dgs} baseline. 
Figure~\ref{fig:nvs_qualitative} shows 2D comparisons of NVS results on two representative scenes, including outputs from 2DGS and our method (GS4B), alongside the ground-truth RGB images. 
While overall image quality was comparable between the methods, GS4B yielded more stable reconstructions in structurally challenging areas. 

\begin{figure}[ht]
    \centering
    \includegraphics[width=\linewidth]{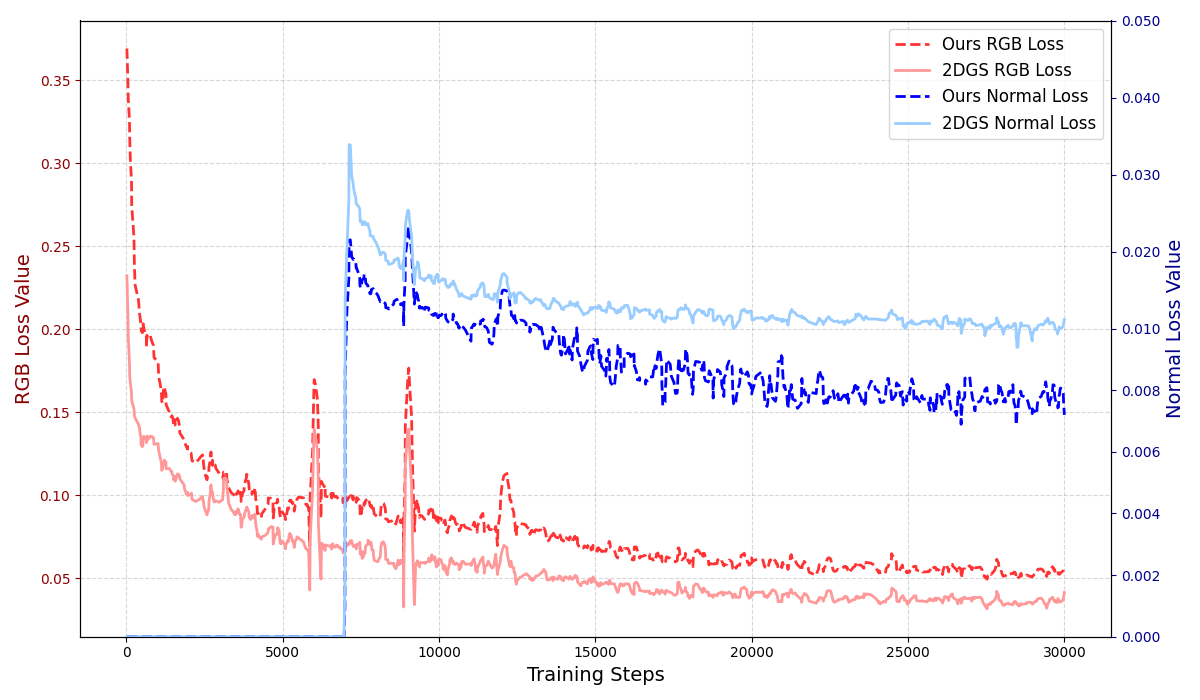}
    \caption{
        Training convergence of our GS4B compared to \gls{2dgs} on a representative scene.
        The plot shows the composite image loss and the scene-wide normal loss over 30k iterations.
    }
    \label{fig:training_curve}
\end{figure}

For quantitative evaluation, Table~\ref{tab:nvs_quantitative} reports the average PSNR, SSIM, and LPIPS scores across the selected scenes. 
Our GS4B method achieved comparable photometric performance to the 2DGS baseline, with slightly higher average PSNR (17.369 vs. 17.190) and SSIM (0.568 vs. 0.552), and a nearly identical LPIPS score (0.260). 
These results confirm that the integration of structural priors does not compromise image fidelity, while preserving perceptual quality across diverse urban settings. 
\begin{table}[ht]
    \centering
    \caption{
        2D quantitative comparison of NVS performance across building scenes from the TUM2TWIN dataset.
        $\uparrow$ indicates higher is better, $\downarrow$ indicates lower is better. 
        Best results per row are highlighted in \colorbox{DarkGreen!50}{\textbf{green}}.
    }
    \label{tab:nvs_quantitative}
    \setlength{\tabcolsep}{3pt}
    \begin{tabular}{l|ccc|ccc}
        \toprule
        \multirow{2}{*}{Scene} & \multicolumn{3}{c|}{2DGS} & \multicolumn{3}{c}{GS4B (ours)} \\
        \cmidrule{2-7}
        & PSNR$\uparrow$ & SSIM$\uparrow$ & LPIPS$\downarrow$ & PSNR$\uparrow$ & SSIM$\uparrow$ & LPIPS$\downarrow$ \\
        \midrule
        1 & \cellcolor{DarkGreen!50}\textbf{19.593} & \cellcolor{DarkGreen!50}\textbf{0.645} & \cellcolor{DarkGreen!50}\textbf{0.222} & 19.224 & 0.610 & 0.274 \\
        2 & 15.785 & 0.434 & 0.355 & \cellcolor{DarkGreen!50}\textbf{15.852} & \cellcolor{DarkGreen!50}\textbf{0.484} & \cellcolor{DarkGreen!50}\textbf{0.342} \\
        3 & 13.520 & 0.298 & 0.348 & \cellcolor{DarkGreen!50}\textbf{14.384} & \cellcolor{DarkGreen!50}\textbf{0.346} & \cellcolor{DarkGreen!50}\textbf{0.306} \\
        4 & \cellcolor{DarkGreen!50}\textbf{19.040} & 0.700 & 0.156 & 19.031 & \cellcolor{DarkGreen!50}\textbf{0.710} & \cellcolor{DarkGreen!50}\textbf{0.155} \\
        5 & 12.332 & 0.271 & 0.482 & \cellcolor{DarkGreen!50}\textbf{13.557} & \cellcolor{DarkGreen!50}\textbf{0.322} & \cellcolor{DarkGreen!50}\textbf{0.425} \\
        6 & \cellcolor{DarkGreen!50}\textbf{13.449} & \cellcolor{DarkGreen!50}\textbf{0.402} & \cellcolor{DarkGreen!50}\textbf{0.365} & 12.870 & 0.347 & 0.414 \\
        7 & \cellcolor{DarkGreen!50}\textbf{19.654} & \cellcolor{DarkGreen!50}\textbf{0.731} & \cellcolor{DarkGreen!50}\textbf{0.135} & 19.514 & 0.701 & 0.146 \\
        8 & \cellcolor{DarkGreen!50}\textbf{21.596} & 0.746 & 0.135 & 21.566 & \cellcolor{DarkGreen!50}\textbf{0.798} & \cellcolor{DarkGreen!50}\textbf{0.143} \\
        9 & 19.737 & 0.742 & 0.150 & \cellcolor{DarkGreen!50}\textbf{20.319} & \cellcolor{DarkGreen!50}\textbf{0.797} & \cellcolor{DarkGreen!50}\textbf{0.138} \\
        \midrule
        Avg. & 17.190 & 0.552 & 0.261 & \cellcolor{DarkGreen!80}\textbf{17.369} & \cellcolor{DarkGreen!80}\textbf{0.568} & \cellcolor{DarkGreen!80}\textbf{0.260} \\
        \bottomrule
    \end{tabular}
\end{table}
\paragraph{Training Convergence}
\begin{sloppypar}
Figure~\ref{fig:training_curve} illustrates the training convergence behavior on a representative scene.
We visualized the composite image loss and the global normal loss over iterations. 
While our method initially exhibited a higher image loss due to its prior-based initialization, it quickly converged to a level comparable with the baseline \gls{2dgs} method.
In terms of normal loss, our method achieved faster convergence and ultimately lower error, indicating improved surface consistency across the entire scene.
\end{sloppypar}
\paragraph{Geometric Accuracy and Completeness}
We evaluated the 3D geometric quality of the reconstructed scenes from two perspectives: accuracy and completeness. 
For geometric accuracy, we compared the reconstructed meshes from \gls{mvs}, \gls{2dgs}, and our method against ground-truth laser-scanned point clouds. 
We used two established 3D geometric metrics: Chamfer Distance (CD), which measures overall geometric similarity, and M3C2, which captures fine-grained surface deviations with local orientation awareness. 
Table~\ref{tab:geometry_accuracy} reports the results across the building scenes from the TUM2TWIN dataset, and shows that our method achieves lower errors than the baselines, indicating improved reconstruction fidelity. 
To assess reconstruction completeness, we used reference point clouds sampled from the \gls{lod}3 building models. 
Unlike laser scans that may be incomplete due to occlusion, the \gls{lod}3-derived data provides full building geometry, making it more suitable for evaluating volumetric coverage. 
As reported in Table~\ref{tab:geo_completeness}, we evaluated completeness using two complementary metrics: 
threshold-based completeness and voxel occupancy completeness (VOC), as introduced in Section~\ref{sec:data_eval}.  
Our method outperforms both baselines across most tested scenes.
\begin{table}[ht!]
    \centering
    \caption{
        3D quantitative comparison of reconstruction accuracy using laser-scanned point clouds as reference. 
        Chamfer Distance (CD$\downarrow$) reflects global geometric similarity, while M3C2 Distance (M3C2$\downarrow$) captures fine-grained surface deviation.
    }
    \label{tab:geometry_accuracy}
    \setlength{\tabcolsep}{4pt}
    \begin{tabular}{l|cc|cc|cc}
        \toprule
        \multirow{2}{*}{Scene} & \multicolumn{2}{c|}{MVS} & \multicolumn{2}{c|}{2DGS} & \multicolumn{2}{c}{GS4B (ours)} \\
        \cmidrule{2-7}
        & CD$\downarrow$ & M3C2$\downarrow$ & CD$\downarrow$ & M3C2$\downarrow$ & CD$\downarrow$ & M3C2$\downarrow$ \\
        \midrule
        1 & 0.834 & \cellcolor{DarkGreen!50}\textbf{0.244} & 0.839 & 0.376 & \cellcolor{DarkGreen!50}\textbf{0.826} & 0.384 \\
        2 & \cellcolor{DarkGreen!50}\textbf{0.561} & 0.012 & 1.145 & 0.008 & 1.013 & \cellcolor{DarkGreen!50}\textbf{0.002} \\
        3 & 2.828 & 0.236 & 1.893 & 0.285 & \cellcolor{DarkGreen!50}\textbf{1.448} & \cellcolor{DarkGreen!50}\textbf{0.125} \\
        4 & \cellcolor{DarkGreen!50}\textbf{0.168} & \cellcolor{DarkGreen!50}\textbf{0.032} & 0.204 & 0.075 & 0.249 & 0.048 \\
        5 & 4.361 & 2.226 & 2.791 & 1.914 & \cellcolor{DarkGreen!50}\textbf{2.340} & \cellcolor{DarkGreen!50}\textbf{1.631} \\
        6 & 1.107 & 0.153 & \cellcolor{DarkGreen!50}\textbf{0.728} & \cellcolor{DarkGreen!50}\textbf{0.059} & 0.857 & 0.161 \\
        7 & 3.567 & 0.015 & 2.955 & 0.662 & \cellcolor{DarkGreen!50}\textbf{2.843} & \cellcolor{DarkGreen!50}\textbf{0.013} \\
        8 & \cellcolor{DarkGreen!50}\textbf{0.397} & 0.014 & 0.758 & 0.107 & 0.527 & \cellcolor{DarkGreen!50}\textbf{0.009} \\
        9 & 2.866 & 0.236 & 2.656 & 0.163 & \cellcolor{DarkGreen!50}\textbf{2.570} & \cellcolor{DarkGreen!50}\textbf{0.076} \\
        \midrule
        Ave. & 1.854 & 0.352 & 1.552 & 0.405 & \cellcolor{DarkGreen!80}\textbf{1.408} & \cellcolor{DarkGreen!80}\textbf{0.272} \\
        \bottomrule
    \end{tabular}
\end{table}

\begin{figure*}[ht]
    \centering
    \begin{minipage}[t]{0.08\linewidth}
        \raggedleft \phantom{\textbf{Method}}  
    \end{minipage}
    \begin{minipage}[t]{0.28\linewidth}
        \centering
        \textbf{Laser Scanning GT}
    \end{minipage}
    \begin{minipage}[t]{0.28\linewidth}
        \centering
        \textbf{\gls{lod}3 GT}
    \end{minipage}

    \vspace{2pt}

    \begin{minipage}[t]{0.08\linewidth}
        \raggedleft \phantom{\textbf{Method}}  
    \end{minipage}
    \begin{minipage}[t]{0.28\linewidth}
        \includegraphics[width=\linewidth]{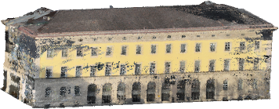}
    \end{minipage}
    \begin{minipage}[t]{0.28\linewidth}
        \includegraphics[width=\linewidth]{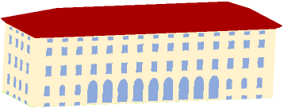}
    \end{minipage}
    
    \vspace{4pt}
    
    \begin{minipage}[t]{0.08\linewidth}
        \raggedleft \textbf{MVS}
    \end{minipage}
    \begin{minipage}[t]{0.28\linewidth}
        \includegraphics[width=\linewidth]{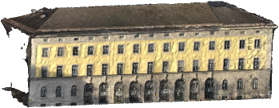}
        \centering Extracted Mesh
    \end{minipage}
    \begin{minipage}[t]{0.28\linewidth}
        \includegraphics[width=\linewidth]{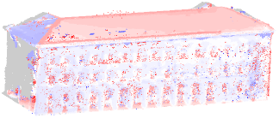}
        \centering M3C2 Distance: 0.244
    \end{minipage}
    \begin{minipage}[t]{0.28\linewidth}
        \includegraphics[width=\linewidth]{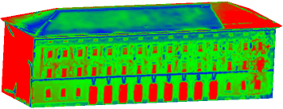}
        \centering Completeness: 0.414
    \end{minipage}

    \begin{minipage}[t]{0.08\linewidth}
        \raggedleft \textbf{2DGS}
    \end{minipage}
    \begin{minipage}[t]{0.28\linewidth}
        \includegraphics[width=\linewidth]{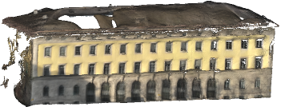}
        \centering Extracted Mesh
    \end{minipage}
    \begin{minipage}[t]{0.28\linewidth}
        \includegraphics[width=\linewidth]{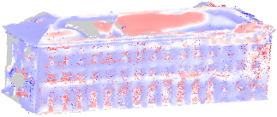}
        \centering M3C2 Distance: 0.376
    \end{minipage}
    \begin{minipage}[t]{0.28\linewidth}
        \includegraphics[width=\linewidth]{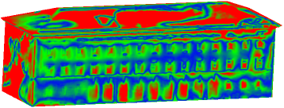}
        \centering Completeness: 0.428
    \end{minipage}

    \begin{minipage}[t]{0.08\linewidth}
        \raggedleft \textbf{GS4B (ours)}
    \end{minipage}
    \begin{minipage}[t]{0.28\linewidth}
        \includegraphics[width=\linewidth]{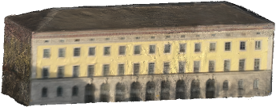}
        \centering Extracted Mesh
    \end{minipage}
    \begin{minipage}[t]{0.28\linewidth}
        \includegraphics[width=\linewidth]{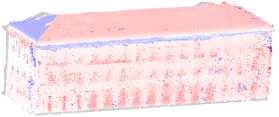}
        \centering M3C2 Distance: 0.384
    \end{minipage}
    \begin{minipage}[t]{0.28\linewidth}
        \includegraphics[width=\linewidth]{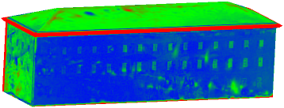}
        \centering Completeness: 0.547
    \end{minipage}

    \begin{minipage}[t]{0.08\linewidth}
        \raggedleft \phantom{\textbf{Method}}  
    \end{minipage}
    \begin{minipage}[t]{0.28\linewidth}
        \includegraphics[width=\linewidth]{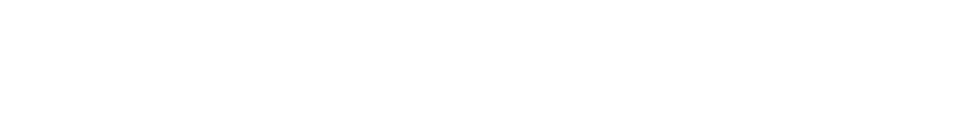}
    \end{minipage}
    \begin{minipage}[t]{0.28\linewidth}
        \includegraphics[width=\linewidth]{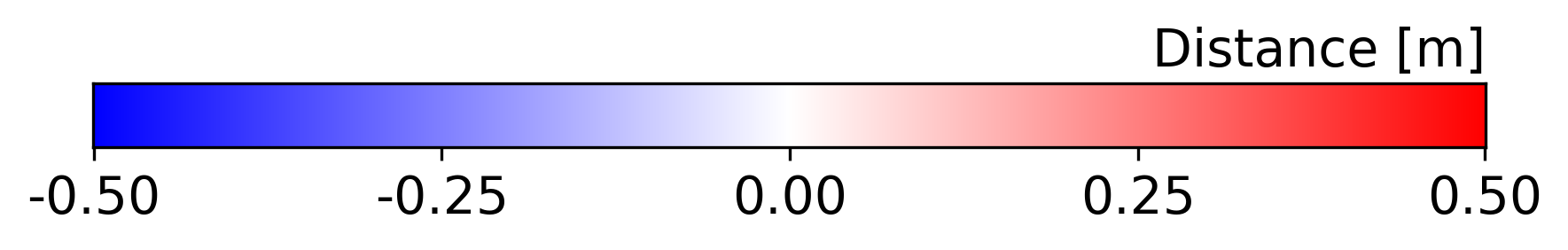}
    \end{minipage}
    \begin{minipage}[t]{0.28\linewidth}
        \includegraphics[width=\linewidth]{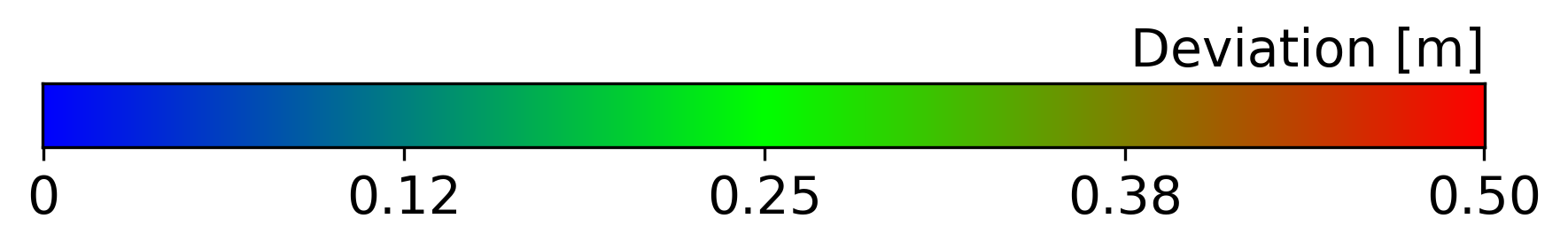}
    \end{minipage}

    \caption{
        Visual comparison of 3D reconstruction accuracy and completeness in a representative urban scene with partial occlusion (i.e., trees, cars). 
        The top row shows ground-truth (GT) point clouds from laser scanning (for accuracy) and \gls{lod}3 building models (for completeness). 
        Rows 2–4 present results from \gls{mvs}, \gls{2dgs}, and our method, including extracted meshes, M3C2 error maps, and completeness visualizations. 
        For completeness, distances from ground-truth points to their nearest reconstructed points are visualized on the GT point cloud using a blue–green–red gradient. 
        Reported values indicate the percentage of points within a 0.5\,m threshold.
    }
    \label{fig:vis_accuracy_completeness}
\end{figure*}

In addition to quantitative comparisons, we provided a visual analysis of reconstruction quality in Figure~\ref{fig:vis_accuracy_completeness}. 
The figure summarizes both geometric accuracy and completeness for a representative urban scene, including occluded regions.
It highlights differences between \gls{mvs}, \gls{2dgs}, and our method GS4B in terms of mesh reconstruction, M3C2 deviation, and completeness distribution. 

\begin{table*}[ht!]
    \centering
    \caption{
        Completeness comparison of 3D reconstruction results using \gls{lod}3-derived point clouds as reference. 
        Threshold-based Completeness reports the percentage of ground truth points recovered within 0.1\,m, 0.2\,m, and 0.5\,m.
        Voxel Occupancy Completeness (VOC) reflects volumetric coverage based on shared voxel occupancy. 
        All values are in $[0, 1]$ and higher is better ($\uparrow$).
    }
    \label{tab:geo_completeness}
    \setlength{\tabcolsep}{6pt}
    \begin{tabular}{l|ccc|c|ccc|c|ccc|c}
        \toprule
        \multirow{2}{*}{Scene} &
        \multicolumn{4}{c|}{MVS} &
        \multicolumn{4}{c|}{2DGS} &
        \multicolumn{4}{c}{GS4B (ours)} \\
        \cmidrule(lr){2-5} \cmidrule(lr){6-9} \cmidrule(lr){10-13}
        & 0.1m$\uparrow$ & 0.2m$\uparrow$ & 0.5m$\uparrow$ & VOC$\uparrow$ 
        & 0.1m$\uparrow$ & 0.2m$\uparrow$ & 0.5m$\uparrow$ & VOC$\uparrow$
        & 0.1m$\uparrow$ & 0.2m$\uparrow$ & 0.5m$\uparrow$ & VOC$\uparrow$ \\
        \midrule
        1 & 0.024 & 0.131 & 0.414 & 0.190 & 0.098 & 0.208 & 0.428 & 0.274 & \cellcolor{DarkGreen!50}\textbf{0.265} & \cellcolor{DarkGreen!50}\textbf{0.387} & \cellcolor{DarkGreen!50}\textbf{0.547} & \cellcolor{DarkGreen!50}\textbf{0.408} \\
        2 & 0.065 & 0.140 & 0.261 & 0.154 & 0.062 & 0.137 & 0.248 & 0.184 & \cellcolor{DarkGreen!50}\textbf{0.066} & \cellcolor{DarkGreen!50}\textbf{0.141} & \cellcolor{DarkGreen!50}\textbf{0.281} & \cellcolor{DarkGreen!50}\textbf{0.188} \\
        3 & 0.012 & 0.034 & 0.096 & 0.051 & \cellcolor{DarkGreen!50}\textbf{0.046} & 0.096 & 0.215 & 0.132 & 0.038 & \cellcolor{DarkGreen!50}\textbf{0.105} & \cellcolor{DarkGreen!50}\textbf{0.273} & \cellcolor{DarkGreen!50}\textbf{0.170} \\
        4 & 0.057 & 0.237 & 0.419 & 0.242 & 0.142 & 0.223 & 0.519 & 0.345 & \cellcolor{DarkGreen!50}\textbf{0.169} & \cellcolor{DarkGreen!50}\textbf{0.322} & \cellcolor{DarkGreen!50}\textbf{0.540} & \cellcolor{DarkGreen!50}\textbf{0.396} \\
        5 & 0.000 & 0.010 & 0.015 & 0.002 & 0.001 & 0.033 & 0.082 & 0.004 & \cellcolor{DarkGreen!50}\textbf{0.006} & \cellcolor{DarkGreen!50}\textbf{0.062} & \cellcolor{DarkGreen!50}\textbf{0.143} & \cellcolor{DarkGreen!50}\textbf{0.012} \\
        6 & 0.102 & 0.208 & 0.472 & 0.281 & \cellcolor{DarkGreen!50}\textbf{0.125} & \cellcolor{DarkGreen!50}\textbf{0.253} & \cellcolor{DarkGreen!50}\textbf{0.551} & 0.314 & 0.110 & 0.241 & 0.472 & \cellcolor{DarkGreen!50}\textbf{0.319} \\
        7 & 0.002 & 0.007 & 0.021 & 0.013 & \cellcolor{DarkGreen!50}\textbf{0.008} & 0.016 & \cellcolor{DarkGreen!50}\textbf{0.045} & 0.025 & 0.004 & \cellcolor{DarkGreen!50}\textbf{0.020} & 0.042 & \cellcolor{DarkGreen!50}\textbf{0.028} \\
        8 & 0.022 & 0.146 & 0.371 & 0.107 & 0.042 & \cellcolor{DarkGreen!50}\textbf{0.199} & 0.398 & 0.167 & \cellcolor{DarkGreen!50}\textbf{0.052} & 0.137 & \cellcolor{DarkGreen!50}\textbf{0.633} & \cellcolor{DarkGreen!50}\textbf{0.245} \\
        9 & 0.078 & 0.171 & 0.334 & 0.187 & 0.076 & 0.163 & 0.374 & 0.221 & \cellcolor{DarkGreen!50}\textbf{0.079} & \cellcolor{DarkGreen!50}\textbf{0.195} & \cellcolor{DarkGreen!50}\textbf{0.394} & \cellcolor{DarkGreen!50}\textbf{0.238} \\
        \midrule
        Avg. & 0.040 & 0.120 & 0.267 & 0.136 & 0.067 & 0.148 & 0.318 & 0.185 & \cellcolor{DarkGreen!80}\textbf{0.088} & \cellcolor{DarkGreen!80}\textbf{0.179} & \cellcolor{DarkGreen!80}\textbf{0.369} & \cellcolor{DarkGreen!80}\textbf{0.223} \\
        \bottomrule
    \end{tabular}
\end{table*}

\paragraph{Ablation Studies }
We conducted ablation studies to evaluate the contribution of three key components in our approach GS4B, considering both NVS and 3D reconstruction quality. 
The evaluated components included: 
(i) the proposed prior-based point cloud initialization, 
(ii) depth prior supervision via the loss term $\mathcal{L}_{\text{d\_b}}$, and 
(iii) normal prior supervision via $\mathcal{L}_{\text{n\_b}}$. 
Table~\ref{tab:ablation_full} summarizes the quantitative results across five representative metrics: PSNR, SSIM, LPIPS for 2D synthesis quality, and Chamfer Distance (CD) and Voxel Occupancy Completeness (VOC) for geometric accuracy and completeness. 
Replacing prior-based initialization with an \gls{sfm}-derived point cloud yielded similar image quality but required more preprocessing. 
Omitting either depth or normal priors resulted in noticeable degradation in 3D reconstruction accuracy and coverage.
\begin{table}[ht!]
    \centering
    \caption{
        Ablation study results on NVS and 3D reconstruction quality.
        Metrics include PSNR (dB), SSIM, and LPIPS for \gls{nvs}; Chamfer Distance (CD) and Voxel Occupancy Completeness (VOC) for 3D geometry.
    }
    \label{tab:ablation_full}
    \setlength{\tabcolsep}{2pt}
    \begin{tabular}{l|ccc|cc}
        \toprule
        Method & PSNR$\uparrow$ & SSIM$\uparrow$ & LPIPS$\downarrow$ & CD$\downarrow$ & VoC$\uparrow$ \\
        \midrule
        With SfM Init         & 19.062 & \cellcolor{DarkGreen!50}\textbf{0.613} & 0.277 & 1.096 & 0.325 \\
        Without Depth Prior   & 18.756 & 0.601 & 0.279 & 2.425 & 0.234 \\
        Without Normal Prior  & 19.154 & 0.609 & 0.276 & 2.017 & 0.395 \\
        \textbf{Ours (Full)}  & \cellcolor{DarkGreen!50}\textbf{19.224} & 0.610 & \cellcolor{DarkGreen!50}\textbf{0.274} & \cellcolor{DarkGreen!50}\textbf{0.826} & \cellcolor{DarkGreen!50}\textbf{0.408} \\
        \bottomrule
    \end{tabular}
\end{table}
\end{sloppypar}
\subsection{Discussion}
\begin{sloppypar}
This section analyzes the performance of GS4B in both 2D and 3D evaluations and concludes with a discussion of its limitations and future prospects.
\paragraph{2D View Synthesis Performance }
We analyze the quality of NVS from both qualitative and quantitative perspectives, complemented by training convergence behavior. 
Figure~\ref{fig:nvs_qualitative} presents visual comparisons across two representative scenes. 
In the first example (Figures~\ref{fig:nvs_qualitative}a–c), where rooftops and walls lack texture, both methods achieve similar overall photometric quality, but our results appear more coherent in these regions, likely due to geometric priors guiding the training. 
In contrast, the second scene (Figures~\ref{fig:nvs_qualitative} d–f) involves sparse viewpoints (only 11 input images), resulting in visibly degraded output from \gls{2dgs}, while our method GS4B retains structural consistency. 
This highlights the benefit of \gls{lod}2-derived priors in supporting synthesis under sparse view coverage. 
However, we also observe that in well-textured scenes with dense views, \gls{2dgs} may retain finer local details, suggesting that our regularization might slightly oversmooth certain regions. 
These trends are further reflected in the quantitative results in Table~\ref{tab:nvs_quantitative}. 
Our method GS4B achieves slightly better overall performance across all metrics. 
Notably, in Scene 5, corresponding to Figures~\ref{fig:nvs_qualitative} d–f, we observe an 18.8\% increase in SSIM, indicating improved preservation of structural details under challenging conditions. 

Beyond the final rendering quality, training dynamics further highlight the behavior of our method. 
Figure~\ref{fig:training_curve} shows the convergence on a scene with sufficient viewpoint coverage, where \gls{2dgs} achieves slightly better overall photometric metrics. 
Our method begins with a higher image loss due to the prior-based initialization, but quickly converges to a comparable level. 
The modestly higher final loss is likely due to reconstruction errors in non-building areas, which are not explicitly guided by our priors. 
In contrast, our global normal loss declines more rapidly and reaches a lower final value, reflecting improved consistency in surface learning. 
These findings are consistent with the visual and quantitative trends discussed earlier. 
\paragraph{3D Reconstruction Analysis }
We evaluate 3D reconstruction quality in terms of both geometric accuracy and completeness. 
As shown in Table~\ref{tab:geometry_accuracy}, our method GS4B outperforms \gls{2dgs} in most scenes and exceeds traditional \gls{mvs} in several cases. 
While \gls{mvs} remains competitive under dense views and rich textures, it degrades in occluded or complex scenarios. 
Notably, GS4B reduces M3C2 error by 32.8\% over \gls{2dgs} and 22.7\% over \gls{mvs}, indicating stronger local geometric consistency. 
As shown in Figure~\ref{fig:vis_accuracy_completeness} (second column), GS4B achieves lower errors along façades and on the heavily occluded left wall of the building. 
Nevertheless, fine-scale structures such as doors and windows remain less accurately reconstructed, likely due to the coarse resolution of the \gls{lod}2 priors. 
The completeness results in Table~\ref{tab:geo_completeness} further highlight the strengths of our GS4B approach. Across all thresholds and in most scenes, GS4B outperforms both \gls{mvs} and \gls{2dgs} in both threshold-based completeness and voxel occupancy completeness (VOC). 
Since many input views cover only partial façades, while the \gls{lod}3 ground truth models represent complete buildings, the absolute completeness values remain relatively low. 
However, our method still achieves significant improvements. 
For instance, in terms of VOC, GS4B shows a 63.9\% increase compared to \gls{mvs}. 

These improvements arise from two main factors. 
First, the \gls{2dgs} representation, with its planar Gaussian splats, provides a more continuous approximation of surfaces compared to point-based \gls{mvs}, which is more susceptible to fragmentation in low-texture or occluded regions. 
This explains why even the standard \gls{2dgs} generally achieves higher completeness than \gls{mvs}.
Second, our method benefits from integrating geometric priors derived from semantic \gls{lod}2 models. 
Unlike methods relying solely on images, which are constrained by occlusions and limited viewpoints, our GS4B framework leverages the volumetric nature of \gls{lod}2 models to provide a complete 3D structure of buildings. 
This enables supervision not only in visible areas but also in regions where image cues are unreliable or entirely missing. 
While images contribute to realistic appearance reconstruction, especially in non-building surroundings such as vegetation and ground surfaces, the inclusion of 3D structural priors allows the geometry of occluded or unseen areas to be reconstructed more faithfully. 

However, the stronger regularization introduced by our priors can, in some cases, attenuate fine-grained geometric details. 
As shown in Figure~\ref{fig:vis_accuracy_completeness}, our method produces smooth and complete surfaces but tends to miss thin structures, which \gls{mvs} is able to capture more accurately when sufficient image observations are available. 
This trade-off reflects the inherent challenge of balancing structural completeness with local geometric precision.
\paragraph{Limitations and Outlook}
Our method performs well in challenging urban reconstruction scenarios, especially under occlusion and sparse viewpoints. 
However, in scenes with dense observations and rich textures, the strong regularization from \gls{lod}2 priors may oversmooth fine details such as eaves, windows, and doors. 
This underscores the need for a more adaptive use of structural priors, such as dynamically weighting their influence based on local scene characteristics. 
While LoD2 building models are becoming increasingly available in many countries, our method still depends on their availability and quality. 
Future work may explore alternative priors, such as more widely accessible \gls{lod}1 models or CAD-based representations, to improve generalizability across diverse urban environments.
\end{sloppypar}
\section{Conclusions}\label{Conclusions}
\begin{sloppypar}
We present GS4Buildings, a Gaussian Splatting framework guided by semantic \gls{lod}2 building models. 
Our method enhances both 2D view synthesis and 3D reconstruction, particularly in challenging scenarios with occlusion and sparse viewpoints.  
Experiments on the TUM2TWIN dataset demonstrate consistent improvements over traditional \gls{mvs} and standard \gls{2dgs}, including a 32.8\% reduction in surface deviation (M3C2) and a 63.9\% increase in voxel occupancy completeness. 
These results highlight the effectiveness of integrating semantic priors into \gls{gs}-based pipelines for scalable and robust urban reconstruction. 
Building on these results, future work could explore dynamic prior weighting and broader prior sources to enhance reconstruction fidelity across diverse scenes, including high-rise and geometrically complex buildings.
Such advancements may also benefit digital city applications, including urban monitoring and digital twin updates.
\end{sloppypar}
\section{Acknowledgements}\label{Acknowledgements}
\begin{sloppypar}
The authors gratefully acknowledge the Professorship of Remote Sensing Applications at TUM for their support in the acquisition and preprocessing of the TUM2TWIN dataset. Special thanks are extended to Katharina Anders and Jiapan Wang for their dedicated contributions. The authors also thank Benjamin Busam for his insightful comments and helpful suggestions that contributed to improving this work.
\end{sloppypar}
{
	\begin{spacing}{1.17}
		\normalsize
		\bibliography{ISPRSguidelines_authors} 
	\end{spacing}
}

\end{document}